\documentclass[conference]{IEEEtran}
\IEEEoverridecommandlockouts
\usepackage{cite}
\usepackage{amsmath,amssymb,amsfonts}
\usepackage{algorithmic}
\usepackage{algorithm}
\usepackage{float}
\usepackage{graphicx}
\usepackage{textcomp}
\usepackage{tabularx}
\usepackage{xcolor}
\usepackage{cite}
\usepackage{amsmath,amssymb,amsfonts}
\usepackage{algorithmic}
\usepackage{graphicx}
\usepackage{textcomp}
\usepackage{xcolor}
\usepackage{amssymb}

\usepackage{booktabs}
\usepackage{subfigure}
\usepackage{multirow}
\usepackage{booktabs}
\usepackage{cite}
\usepackage{amsmath,amssymb,amsfonts}
\usepackage{algorithmic}
\usepackage{graphicx}
\usepackage{textcomp}
\usepackage{xcolor}
\usepackage{amssymb}
\usepackage{algorithm, multirow}
\usepackage{booktabs}
\usepackage{subfigure}
\usepackage{amsmath,amssymb,amsfonts}
\usepackage{algorithmic}
\usepackage{graphicx}
\usepackage{textcomp}
\usepackage{xcolor}
\usepackage{amssymb}
\usepackage{algorithm, multirow}
\usepackage{booktabs}
\usepackage{subfigure}
\usepackage{caption}
\usepackage{float} 
\usepackage{subcaption}
\usepackage{amsmath}

\usepackage{amsmath}        
\usepackage{amssymb}        
\usepackage{float}          
\usepackage{graphicx}       

\usepackage{cite}
\usepackage{amsmath,amssymb,amsfonts}
\usepackage{algorithmic}
\usepackage{graphicx}
\usepackage{textcomp}
\usepackage{xcolor}

\def\BibTeX{{\rm B\kern-.05em{\sc i\kern-.025em b}\kern-.08em
    T\kern-.1667em\lower.7ex\hbox{E}\kern-.125emX}}

\begin{document}

\title{User to Video: A Model for Spammer Detection Inspired by Video Classification Technology}

\author{
	\IEEEauthorblockN{1\textsuperscript{st} Haoyang Zhang}
	\IEEEauthorblockA{
		International College of CQUPT\\
		Chongqing University of Posts and\\
		Telecommunications\\
		Chongqing, China \\
		2021214892@stu.cqupt.edu.cn}
	\and
	
	\IEEEauthorblockN{2\textsuperscript{nd} Zhou Yang$^*$}
	\IEEEauthorblockA{
		School of Electronics and Internet of\\
		Things\\
		Sichuan Vocational and Technical College\\
		Suining, China \\
		yzhoul392@gmail.com}
		
	\and
	\IEEEauthorblockN{3\textsuperscript{rd} Yucai Pang}
	\IEEEauthorblockA{
		School of Electronics and Internet of\\
		Things\\
		Sichuan Vocational and Technical College\\
		Suining, China \\
		pangyc@cqupt.edu.cn}

\thanks{$^*$ Corresponding author}
}

\maketitle

\begin{abstract}
	This article is inspired by video classification technology. If the user behavior subspace is viewed as a frame image, consecutive frame images are viewed as a video. Following this novel idea, a model for \textbf{s}pammer \textbf{d}etection based on \textbf{u}ser \textbf{v}ideoization, called UVSD, is proposed. Firstly, a user2piexl algorithm for user pixelization is proposed. Considering the adversarial behavior of user stances, the user is viewed as a pixel, and the stance is quantified as the pixel's RGB. Secondly, a behavior2image algorithm is proposed for transforming user behavior subspace into frame images. Low-rank dense vectorization of subspace user relations is performed using representation learning, while cutting and diffusion algorithms are introduced to complete the frame imageization. Finally, user behavior videos are constructed based on temporal features. Subsequently, a video classification algorithm is combined to identify the spammers. Experiments using publicly available datasets, i.e., WEIBO and TWITTER, show an advantage of the UVSD model over state-of-the-art methods.
\end{abstract}


\section{Introduction}
\par As computer technology advances, users can share information on social media platforms at zero cost. With the rise of Large Language Modeling technology, the way spammers manipulate machine users has become even more undetectable. Currently, major social platforms have launched open communities for spammers. Not only in industry, academic sessions are also focusing on this complex task. Top conferences and journals have published research results \cite{traganitis2021identifying,Deng2023Markov,hu2024f2gnn,wu2020graph}. Therefore, it is essential to recognize spammers.
\par There are user relationships in social networks, so existing mainstream strategies are based on graph models for modeling historical behaviors. For instance, GNN is widely used as the final modeling component\cite{Deng2023Markov,yi2023spammer,wu2024graph,liu2024disentangled}. However, as the number of historical user behaviors increases, the social spreading graph is too large, thus requiring many hardware resources. To address this problem, this work proposes a new view to model historical behaviors. In the video domain, individual frame images contain many pixel points. Moreover, there are high-intensity relational features of neighboring pixel points. Inspired by video modeling techniques, if individual elements of the historical behavioral spread space are considered pixel points, individual subspaces can be viewed as frame images. As a result, the entire spread space modeling task can then be transformed into a video modeling task. Based on the above ideas, this article proposes a social spammer identification model based on the videoization strategy for user behaviors. The main contributions are as follows:
\begin{itemize}
	\setlength{\itemsep}{-1pt}
	\setlength{\parsep}{0pt}
	\setlength{\parskip}{0pt}
	\vspace{-2mm}
	\item[1.]
	Subspace Imageization. User behavior graphs are non-Euclidean structures. Therefore, the high-dimensional graph is transformed into a two-dimensional space with the help of representation learning. Subsequently, cutting and diffusion algorithms are introduced to transform the subspace into frame images.
	\item[2.]	
	Space Videoization. It is time-slicing the entire behavioral space. Meanwhile, the entire space is transformed into a set of frame images with the help of behavior2image and user2pixel algorithms. Subsequently, user videos are spliced based on temporal features.
	\item[3.] To my knowledge, the idea of videoing historical behavior is presented for the first time. Moreover, it has been implemented. Similarly, the effectiveness of the pixelization strategy has been tested in other tasks\cite{Xiao2022Diffusion,Pang2023Topic}. In particular, it is different with the above algorithms.
\end{itemize}

\section{Related Work}
\par To solve the problem of spammer detection, researchers have modeled user behavior features. For instance, based on graph attention mechanism model \cite{agarwal2022modeling}. They construct a graph based on user relationships and then identify fraudulent review users. The above method provides a feature representation of the user-relationship graph, but ignore the importance of heterogeneous relationships between user-entitity. Later, heterogeneous user-relationship graph based on user-entity were proposed \cite{ZHANG2023Detecting}. They used graph model for modeling user behaviors to identify spammers. All of the above approaches are based on graph models, but ignore the fact that graph models constructed based on a large number of user behaviors consume a large amount of memory and arithmetic power.
\par Inspired by the video classification technology, a spammer detection model based on user videoization is proposed. Similar methodological possibilities have been demonstrated. For example, a rumor detection model based on topic imageization \cite{Pang2023Topic}. They transform the topic propagation graph into images and use convolutional neural networks to identify rumors. Our work proposes a new view to model historical user behavior. By considering individual elements in the historical behavior diffusion space as pixel points, the individual subspaces can be considered as frame images, which enables the conversion of the entire expansion subspace modeling task into a video modeling task.

\section{Methodology}
\subsection{Problem Formalization}
\par Define $\xi_i \rightarrow \widehat{y_{i}}$ denotes the UVSD model. $\xi_i=<E_i, F_i>$ denote the historical behavior graph of the $i$th user $u_i$. $E_i$ denotes the set of relationships of user nodes, and $F_i$ denotes node features. The $\widehat{y_{i}}$ denotes the prediction, i.e. $\widehat{y_{i}} \in [0, 1]$.
\par
\begin{figure}[h]
	\center{\includegraphics[width=8.7cm] {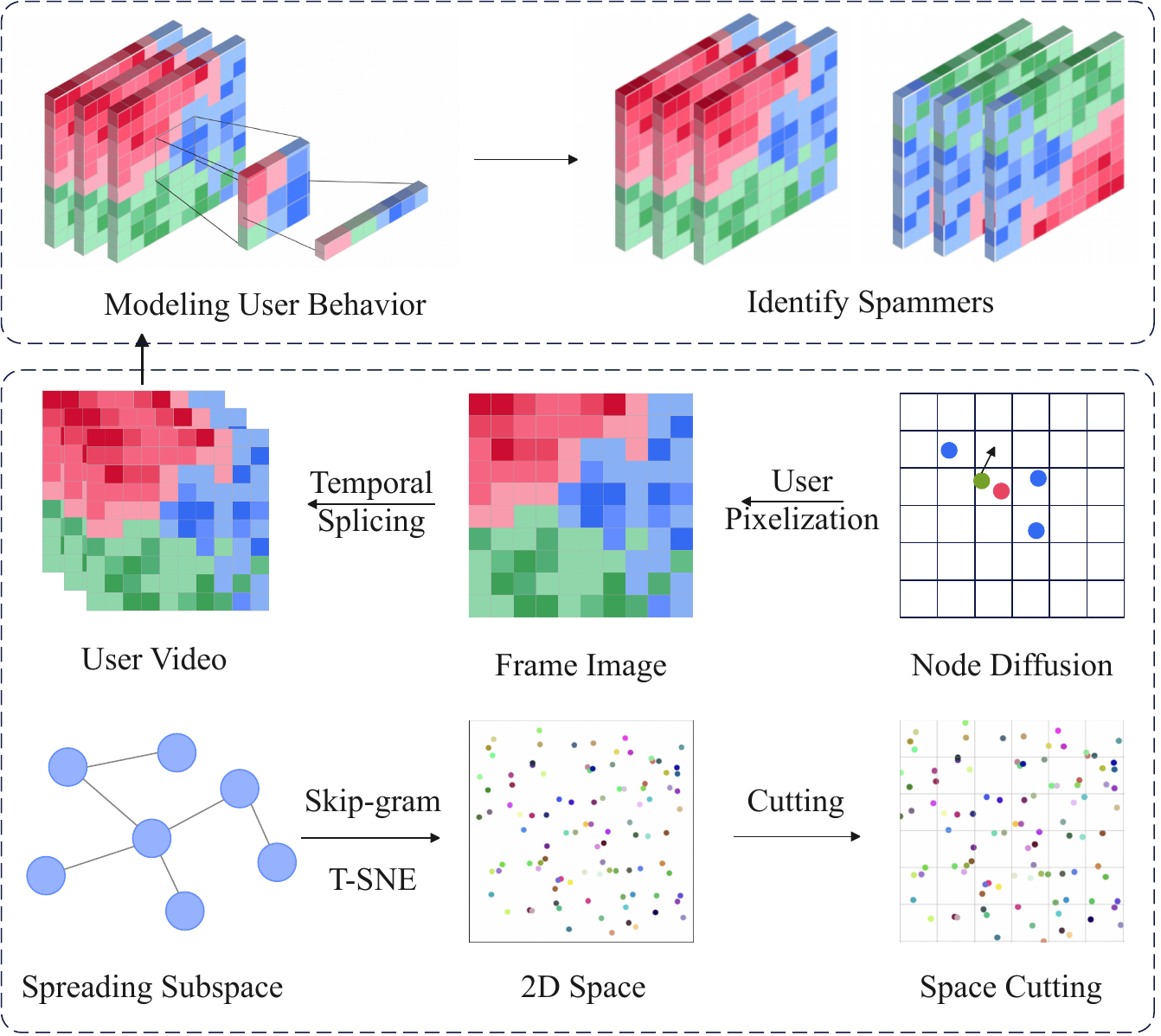}} 
	\caption{Framework of the UVSD model.}
	\label{fig-model}
\end{figure}

 Inspired by video classification technology, a spammer detection model using user videoization is proposed (see Fig. \ref{fig-model}). The model contains four steps: user pixelization, subspace imageization, and video-based spammer identification.
\subsection{User Pixelization}
\par Spammers usually post war-inducing statements to manipulate the direction of public opinion. Therefore, in this paper, the user stance feature is viewed as the RGB value of the pixel point. The user2pixel formula is shown below:
\begin{equation}
	pixel^{rgb}_{i}=
	\begin{cases}
		red, \ if\ emo_{i}=PosStance\\
		green, \ if\ emo_{i}=negStance\\
		blue, \ if\ emo_{i}=neuStance
	\end{cases}
\end{equation}
where $pixel^{rgb}_{i}$ denotes the RGB value of pixel point $pixel_i$ for user $u_i$. $emo_{i}$ denotes the user's stance quantified using the pre-trained neural network, including positive, negative, and neutral stances. Subsequently, the comprehensive user influence is mapped to the brightness value of a pixel point. The specific formula is shown below:
\begin{equation}
	piexl^{bri}_i= officLev_i+\psi \times fansLev_i
\end{equation}
where $piexl^{bri}_i$ denotes the brightness value of pixel $pixel_i$. $\eta$ denotes the base value. $fansLev_i=\frac{fansNum_i}{fansThre}$ denotes the fans level. $fansNum_i$ denotes the fans' number of user $u_i$. $fansThre$ denotes the threshold between fans levels. $officLev_i$ denotes the official level. As officers usually act in high-influence accounts, an attenuation factor $\psi$ is introduced to reduce the influence of star accounts, $\psi \in [0, 1]$.
\subsection{Subspace Frame Imageization}
\par This section proposes an algorithm called behavior2image. The algorithm contains three steps: node representation, feature representation, and frame imageization.
\par \textbf{Node Representation:} Using the Node2vec algorithm, the nodes of the $j$th subgraph $\xi_{ij}$ of the historical behavior are embedded in a two-dimensional space. Let user $u_0$ denote the source node. Subsequently, the formula for selecting the next node is shown below:
\begin{equation}
	P(u_i=\mu \mid u_{i - 1}=\nu)=\left\{
	\begin{aligned}
		{\frac{\varphi_{\mu \nu}}{Z}, (\mu ,\nu)\in E_{ij}}\\
		{0\ \ , (\mu ,\nu)\notin E_{ij}}\\
	\end{aligned}
	\right.
\end{equation}
where $Z$ denotes the regularization factor. $\varphi_{\mu \nu}$ denotes the transfer probability from node $\nu$ to node $\mu$, i.e., $\varphi_{\mu \nu}$ is the weight of the edge $(\mu, \nu)$ $\varpi_{\mu \nu}$. Therefore, there is exists $\varphi_{\mu \nu}=\varpi_{\mu \nu }$. The random walk of the second-order graph improves $\varphi_{\mu \nu}=\varpi_{\mu \nu}$ to $\varphi_{\mu \nu}=\lambda _{p q}(t, \nu)\cdot \varpi_{\mu \nu}$. where the formula for $\lambda _{p q}(t, \nu)$ is shown below:
\begin{equation}
	\lambda _{p q}(t, \nu)=\left\{
	\begin{aligned}
		{1/p, d_{t, \nu}=0}\\
		{1, d_{t, \nu}=1}\\
		{1/q, d_{t, \nu}=2}\\
	\end{aligned}
	\right.
\end{equation}
The parameters $p$ and $q$ are introduced to balance the DFS (depth-first search) and BFS (breadth-first search) strategy probabilities, i.e., $p=q=0.5$.
\par \textbf{Feature Representation:} Combined with the feature learning algorithm, Skip-gram maps the nodes to a high-dimensional space while ensuring solid relationships in the subgraph $\xi_{ij}$. Subsequently, the user's high-dimensional representation vectors are downscaled by the nonlinear dimensionality reduction algorithm T-SNE, so that the subgraph $\xi_{ij}$ is transformed into a two-dimensional space.
\par \textbf{Frame Imageization:} The high dimensional subspace $\xi_{ij}$ has been transformed into a two dimensional space. However, the user nodes in the subspace are unordered. Therefore, the space is first grid cut. Subsequently, the nodes are sequentially pixel-aligned. The pixel grid cut distance computes formula is shown below:
\begin{equation}
	cut^{dis}_{ij}=\sqrt{\frac{h^{dis}_{ij}\cdot v^{dis}_{ij}}{\Delta n\cdot \gamma}}
\end{equation}
where $cut^{dis}_{ij}$ denotes the pixel grid length of the subspace $\xi_{ij}$. $h^{dis}_{ij}$ and $v^{dis}_{ij}$ denote the distance difference between the horizontal and vertical directions of the subspace. $\Delta n$ denotes the node number in the subspace $\xi_{ij}$. $\gamma$ denotes the number of pixel grids assigned to the nodes. Localized areas exist where user relationships are stronger because of the different user relationships in the subspace. If a pixel grid is assigned to each user node, it will result in missing node relationships. Meanwhile, it will also increase the time spent on model computing.
\begin{figure}[h]
	\center{\includegraphics[width=8.7cm] {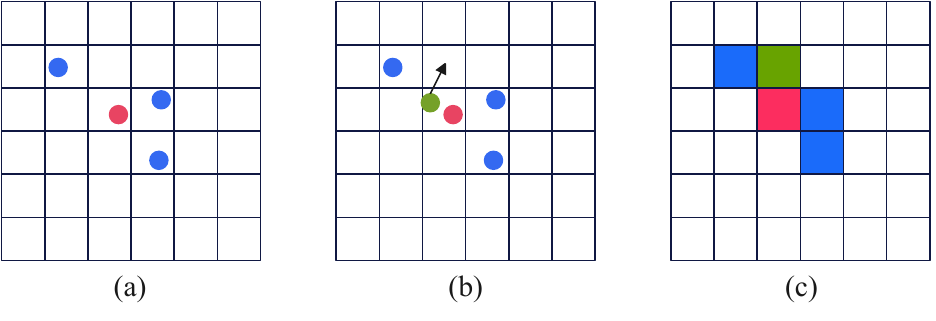}} 
	\caption{Cutting and diffusion algorithm.}
	\label{fig-cut}
\end{figure}
\par After the cutting algorithm, multiple user nodes in a single-pixel grid are unavoidable. Therefore, the diffusion algorithm is introduced to image the subspace. The algorithmic procedure is shown as follows: 1) Only a single user node exists in the pixel grid scope (see Fig. \ref{fig-cut}(a)). In this case, the RGB of the pixel point is $pixel^{rgb}_{i}$, while the brightness is $piexl^{bri}_i$. 2) Multiple user nodes exist (see Fig. \ref{fig-cut}(b)). One user node is randomly retained, and other nodes are spread to the neighboring pixel grid in the clockwise direction.
\subsection{Video Construction and Spammer Detection}
\par The user behavior graph $\xi_i$ is time-sliced to obtain a sequence of user behavior subgraphs $\widehat{\xi_{i}}=\{\xi_{i1}, \xi_{i2}, ... , \xi_{in}\}$. Due to the different number of user behaviors at different times, i.e., active during the day and sparse at night. Therefore, the slicing strategy based on temporal features leads to inconsistent sizes of images in different frames. To solve this problem, this paper adopts a temporal slicing strategy with a uniform number of users. The number of subgraph users in the sequence $\widehat{\xi_{i}}$ is an equal-variance series, and the difference is $\Delta n$.
\par However, the number of user nodes in any subgraph in the subgraph sequence $\widehat{\xi_{i}}$ is still different. For this reason, the user nodes belonging to $\xi_{i(k-1)}$ in the subgraph $\xi_{ik}$ are deleted. Thus, the number of nodes in the subgraph sequence $\widehat{\xi_{i}}$ is normalized. The specific formula is shown below:
\begin{equation}
	\widehat{\xi_{ik}}=\left\{
	\begin{aligned}
		{\xi_{ik} \circleddash \xi_{i(k-1)}, \ if \ k>1}\\
		{\xi_{ik}\ \ \ \ \ \ \ \ \ \ ,  \  if \ k=1} \\
	\end{aligned}
	\right.
\end{equation}
where $\widehat{\xi_{ik}}$ denotes the normalized subgraph of $\xi_{ik}$. $\xi_{i(k-1)}$ denotes the previous subgraph of $\xi_{ik}$ and $\xi_{i(k-1)}$ is a subgraph of $\xi_{ik}$, i.e., $\xi_{i(k-1)} \in \xi_{ik}$.
\par Let $\pi_{\mu \nu}$ denote the relationship weights between nodes in the subgraph $\xi_{ik}$. Subsequently, the relationship weight $\pi_{\mu \nu}$ of the subgraph $\widehat{\xi_{ik}}$ is updated to $\widehat{\pi_{\mu \nu}}$. The specific formula is shown below:
\begin{equation}
	\widehat{\pi_{\mu \nu}}=\frac{\sum^{\nu}_{l=\mu}\pi_{l(l+1)}}{\sum^{\nu}_{l=\mu} +1}
\end{equation}
where $\sum^{\nu}_{l=\mu}\pi_{l(l+1)}$ denotes the sum of the weights of the paths from node $\mu$ to $\nu$ through the deleted node. The $\sum^{\nu}_{l=\mu}$ denotes the number of nodes in the path that belongs to the subgraph $\xi_{i(k-1)}$.
\par Subsequently, the behavior2image algorithm is used to transform the normalized subgraph sequence $\{\widehat{\xi_{i1}}, \widehat{\xi_{i2}}, ... , \widehat{\xi_{in}}\}$ into a set of frame images $\{image_{i1}, image_{i2}, ... , image_{in}\}$. Meanwhile, it is connected as a user video based on temporal features. The specific formula is shown below:
\begin{equation}
	video_{i}=\text{concat}(image_{i1}, image_{i2}, ..., image_{in})
\end{equation}
\begin{equation}
	\widehat{y_{i}}=cnn\_model(video_{i})
\end{equation}
where $video_{i}$ denotes the video from user $u_i$. $y_{i}$ denotes the UVSD model prediction.

\subsection{Learning Algorithm and Time Complexity}
\begin{algorithm}
	\caption{UVSD Model}
	
	\textbf{Input:} Historical behavior graph $\xi_i = \langle E_i, F_i \rangle$ of user $u_i$ \\
	\textbf{Output:} Prediction $\widehat{{y}_i} \in [0,1]$
	
	\begin{algorithmic}[1]
		\STATE \textbf{User Pixelization:}
		\FOR{each user $u_i$ in $\xi_i$}
		\STATE Quantify stance $emo_i$ using pre-trained neural network
		\STATE Assign RGB value: $pixel^{rgb}_i \gets$ Eq. (1)
		\STATE Compute brightness: $pixel^{bri}_i \gets officLev_i + \psi \times fansLev_i$ (Eq. 2)
		\ENDFOR
		
		\STATE \textbf{Subspace Imageization:}
		\FOR{each subgraph $\xi_{ij}$ in time-sliced sequence $\widehat{\xi}_i$}
		\STATE \textbf{Node Representation:}
		\STATE Embed nodes via Node2vec with transition probability $P(u_i=\mu \mid u_{i-1}=\nu)$ (Eq. 3)
		\STATE Adjust $\varphi_{\mu\nu}$ using $\lambda_{pq}(t,\nu)$ (Eq. 4)
		\STATE \textbf{Feature Representation:}
		\STATE Apply Skip-gram and T-SNE to reduce node embeddings to 2D space
		\STATE \textbf{Frame Imageization:}
		\STATE Grid-cut subspace $\xi_{ij}$ with $cut^{dis}_{ij} \gets \sqrt{\frac{h^{dis}_{ij} \cdot v^{dis}_{ij}}{\Delta n \cdot \gamma}}$ (Eq. 5)
		\STATE Diffuse overlapping nodes clockwise (Fig. 2(b))
		\STATE Assign pixel RGB/brightness based on $pixel^{rgb}_i$ and $pixel^{bri}_i$
		\ENDFOR
		
		\STATE \textbf{Video Construction:}
		\STATE Normalize subgraph sequence by (Eq. 6)
		\STATE Update edge weights $\widehat{\pi_{\mu\nu}} \gets \frac{\sum_{l=\mu}^{\nu} \pi_{l(l+1)}}{\sum_{l=\mu}^{\nu} + 1}$ (Eq. 7)
		\STATE Generate frame images $\{image_{i1},...,image_{in}\}$ via behavior2image
		\STATE Concatenate frames by (Eq. 8)
		
		\STATE \textbf{Spammer Detection:}
		\STATE Predict $\widehat{{y}_i} \gets \text{cnn\_model}(video_i)$ (Eq. 9)
		
		\STATE \textbf{return} $\widehat{{y}_i}$
	\end{algorithmic}
\end{algorithm}

\begin{table*}[h!]
	\centering
	\scalebox{1.0}{
		\begin{tabular}{c*{14}{c}}
			\toprule
			\multirow{2}{*}{Method} & \multicolumn{2}{c}{5\%} & \multicolumn{2}{c}{10\%} & \multicolumn{2}{c}{25\%} & \multicolumn{2}{c}{50\%} & \multicolumn{2}{c}{75\%} & \multicolumn{2}{c}{100\%} \\
			\cmidrule(lr){2-3} \cmidrule(lr){4-5} \cmidrule(lr){6-7} \cmidrule(lr){8-9} \cmidrule(lr){10-11} \cmidrule(lr){12-13}
			& Acc. & AUC & Acc. & AUC & Acc. & AUC & Acc. & AUC & Acc. & AUC & Acc. & AUC \\
			\midrule
			SMSFR      & 0.504 & \boxed{0.511} & \boxed{0.513} & \boxed{0.517} & \boxed{0.518} & \boxed{0.525} & \boxed{0.536} & \boxed{0.523} & \boxed{0.534} & \boxed{0.579} & \boxed{0.551} & \boxed{0.540} \\
			OSSD       & 0.529 & 0.542 & 0.536 & 0.567 & 0.594 & 0.629 & 0.616 & 0.627 & 0.623 & 0.655 & 0.632 & 0.658 \\
			SSDMV      & \boxed{0.474} & 0.521 & 0.553 & 0.583 & 0.617 & 0.642 & 0.688 & 0.724 & 0.701 & 0.733 & 0.703 & 0.756 \\
			GCNwithMRF & 0.643 & 0.702 & 0.726 & 0.785 & 0.774 & 0.815 & 0.801 & 0.835 & 0.811 & 0.855 & 0.812 & 0.857 \\
			Nash-Detect& 0.651 & 0.723 & 0.724 & 0.794 & 0.785 & 0.845 & 0.789 & 0.828 & 0.805 & 0.844 & 0.809 & 0.852 \\
			MDGCN      & \underline{0.733} & \textbf{0.802} & 0.794 & \underline{0.845} & \underline{0.822} & \underline{0.888} & 0.814 & 0.872 & 0.818 & 0.869 & 0.817 & 0.887 \\
			SEINE      & 0.705 & 0.755 & \underline{0.798} & 0.833 & 0.817 & 0.878 & \underline{0.822} & \underline{0.882} & \underline{0.825} & \underline{0.874} & \underline{0.827} & \underline{0.891} \\
			UVSD & \textbf{0.735} & \underline{0.781} & \textbf{0.812} & \textbf{0.858} & \textbf{0.824} & \textbf{0.894} & \textbf{0.836} & \textbf{0.887} & \textbf{0.833} & \textbf{0.886} & \textbf{0.835} & \textbf{0.896} \\
			\bottomrule
		\end{tabular}
	}
	\caption{Spammer Detection on WEIBO dataset. To validate the most effective length of time for historical user behaviors, various Spread Behavior Percentages (SBP) are used for performance comparison. When SBP=100\%, it indicates that the user’s data for the recent half-year is input into the model for training or prediction.}
	\label{tab:performance1}
\end{table*}

\par For details of the pseudo code algorithm, see Table \textbf{Algorithm 1} UVSD Model.
\par Let $N_1$ and $N_2$ represent the number of user nodes and the number of temporal slices, respectively. $M$ represents the number of user relationships in a single temporal slice. The component of user pixelization is mostly done in the pre-processing phase, and the time complexity is $O(N_1 M)$. The subspace frame imageization component mainly employs the Node2vec algorithm and T-SNE for dimensionality reduction, and thus the time consumption is $O(N_1^2)$. The video construction and spammer detection component uses a CNN model, and the time complexity is $O(N_2×N_1^2)$. In summary, the overall time complexity of the UVSD model is $O(N_1 M+N_1^2+N_2×N_1^2)$.

\section{Experiments and Analysis}
\subsection{Datasets and Settings}
\par \textbf{Datasets:} We validate performance on publicly available datasets, WEIBO and TWITTER. In the TWITTER dataset, Yang et al.\cite{yang2012analyzing} collected $487$ spammers and $8770$ non-spammers. Subsequently, we construct the WEIBO dataset\footnote{https://github.com/yzhouli/Spammer-Detection-Dataset} from rumor datasets\cite{Song2021CED,ma2016detecting,Yang2024model,xu2023contrastive} and WEIBO community management center. We selected $342$ spammers and $343$ non-spammers. The statistics of the datasets are shown in table \ref{table_dataset}.
\begin{table}[!h]
	\renewcommand{\arraystretch}{1.5}
	\centering
	\scalebox{0.99}{	
		\begin{tabular}{c|c c}\toprule[1.3pt]
			Statistic& TWITTER & WEIBO  \\ \hline \hline
			$\#$ of Respones & $224280712$  & $1473684910$ \\ \hline
			$\#$ of Events& $950342$ & $1436896$  \\ \hline
			$\#$ of Users& $5657$ & $685$  \\ \hline
			$\#$ of Non-Spammers& $8770$ & $343$  \\ \hline
			$\#$ of Spammers& $487$ & $342$  \\ \hline \hline
			\toprule[1.3pt]
		\end{tabular}
	}
	\caption{STATISTICS OF THE DATASETS}
	\label{table_dataset}
\end{table}

\par \textbf{Settings:} The model and baseline algorithm are validated on two publicly available datasets. We divided the data into training and test sets at a ratio of 8:2. Subsequently, Adam is chosen as the model optimizer, and the learning rate is set to 0.001. Meanwhile, the training time of the model is set to 100 epochs, and the optimal model parameters are preserved using an early stopping mechanism. The model's hyper-parameters are set:$fansThre$=1000, $\psi$=0.62/0.68, $\Delta n=6400$, and $\gamma=1.3$. Finally, the parameter settings for the baseline algorithm are determined from the original article.
\subsection{Baseline Algorithms}
\par We compared the UVSD model with the baseline algorithm. \textbf{SMSFR:} A user representation model using a matrix factorization mechanism \cite{zhu2012discovering}. \textbf{OSSD:} A spammer identification model \cite{hu2014social} introducing social behavior directionality to improve SMSFR\cite{zhu2012discovering}. \textbf{SSDMV:} A semi-supervised spammer detection model based on autoencoder framework\cite{li2018ssdmv}. \textbf{GCNwithMRF:} A spammer identification model combining Markov Random Fields (MRF) and Graph Convolutional Neural Networks (GCN) \cite{wu2020graph}.  \textbf{Nash-Detect:} A robot spammer recognition model based on Nash reinforcement learning\cite{dou2020robust}. \textbf{MDGCN:} A Markov-driven GCN model \cite{Deng2023Markov}. MRF is transformed into ARMRF (Adaptive Reward Markov Random Fields). \textbf{SEINE:} A spammer recognition model based on user interaction and graph neural network (GNN) \cite{Agarwal2022Model}.

\subsection{Evaluation Metrics}
\par The task of spammer detection is a typical binary classification problem. Therefore, the accuracy (Acc.) is used as an evaluation metric for model performance \cite{Ma2023Improving}. The highly unbalanced TWITTER dataset has all baseline method accuracies exceeding 95\% \cite{Deng2023Markov}. Therefore, the models trained on the TWITTER dataset are evaluated using the AUC metric \cite{wu2020graph}.

\subsection{Detection Performance}
\par  To validate the model performance, a comparison of stateof-the-art models for spammer detection is performed. Subsequently, the highly unbalanced TWITTER dataset is evaluated using only the AUC metric\cite{wu2020graph}.

\begin{table}[!h]
	\centering
	\scalebox{0.9}{	
		\begin{tabular}{c|c c c c c c}\toprule[1.0pt]
			
			Methods      & $5\%$ & $10\%$ & $25\%$ & $50\%$ & $75\%$ & $100\%$                                    \\ \hline \hline
			
			SMSFR        & $\boxed{0.059}$  & $\boxed{0.112}$ & $\boxed{0.138}$ & $\boxed{0.157}$ & $\boxed{0.149}$ & $\boxed{0.161}$                        \\ \hline
			OSSD         & $0.114$  & $0.137$ & $0.178$ & $0.187$ & $0.195$ & $0.201$                                   \\ \hline
			SSDMV        & $0.153$  & $0.198$ & $0.257$ & $0.277$ & $0.269$ & $0.278$                                    \\ \hline
			GCNwithMRF   & $0.621$  & $0.694$ & $0.798$ & $0.856$ & $0.853$ & $0.868$                       \\ \hline
			Nash-Detect  & $0.611$  & $0.695$ & $0.819$ & $0.865$ & $0.879$ & $0.886$                     \\ \hline  
			SEINE        & $0.677$  & $0.743$ & $0.882$ & $0.911$ & $0.901$ & $0.915$                   	\\ \hline
			MDGCN        & $\underline{0.685}$  & $\underline{0.778}$ & $\underline{0.888}$ & $\mathbf{0.919}$ & $\underline{0.925}$ & $\underline{0.926}$                   	\\ \hline 
			
			UVSD   & $\mathbf{0.703}$  & $\mathbf{0.797}$ & $\mathbf{0.904}$ & $\underline{0.917}$ & $\mathbf{0.928}$ & $\mathbf{0.931}$              \\ \hline
			\toprule[1.0pt]
		\end{tabular}
	}
	\caption{Spammer Detection on TWITTER dataset.}
	\vspace{-3mm}
	\label{table_dataset1}
\end{table}

\par It can be found that the performance of the traditional baseline models is average, including the matrix factorization and markov algorithms (see Table \ref{tab:performance1} and \ref{table_dataset1} (top)). Compared to the UVSD model, SMSFR, OSSD, SSDMV, and GCNwithMRF improve performance as follows: 1) WEIBO pre-training TOP-1 Acc. and AUC metrics improve by +2.3/3.9\%. 2) TWITTER pre-training TOP-1 AUC metric improves by +6.3\%.
\par The recent deep learning-based, i.e., Nash-Detect, MDGCN and SEINE, model is more effective compared to the traditional algorithms, with best performance differences of +3.4/5.8\% for TOP-1 AUC metrics  (see Table \ref{tab:performance1} and \ref{table_dataset1} (bottom)). This is because the representation capability of deep learning is better than traditional models. Finally, compared to the UVSD model, the neural network model's TOP-1 AUC metrics performance gains are Nash-Detect (+4.4/4.5\%), MDGCN (+0.9/1.6\%) and SEINE (+0.5/0.5\%). For TOP-1 Acc. metrics performance gains are Nash-Detect (+2.6\%), MDGCN (+1.8\%) and SEINE (+0.8\%). Thus, the effectiveness is further demonstrated.

\subsection{Validity Analysis for Videoization Component}
\par Spreading space videoization is based on the transformation of the subspace into frame images using user relations. Therefore, this section verifies the validity using different user relations. \textbf{-w Random:} using random user relations. \textbf{-w Friend:} using the friend relations of the participating users. \textbf{-w Comment:} using the comment relations between users.
\begin{figure}[H]		
	\centering 
	\subfigure[User Relations]{
		\begin{minipage}[t]{0.5\linewidth}
			\centering
			\includegraphics[width=1\linewidth]{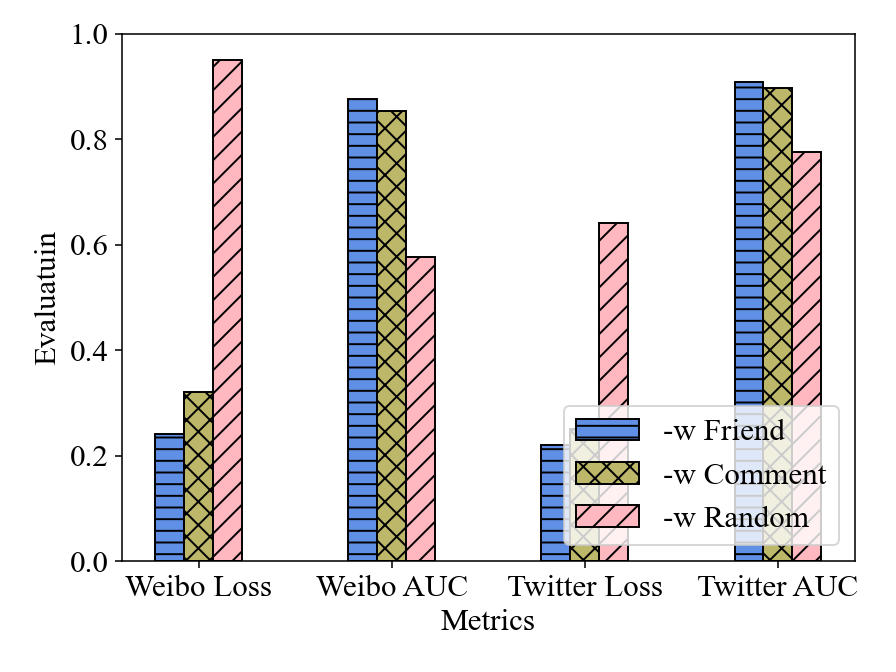}
		\end{minipage}%
	}%
	\subfigure[Relation Combinations]{
		\begin{minipage}[t]{0.5\linewidth}
			\centering
			\includegraphics[width=1\linewidth]{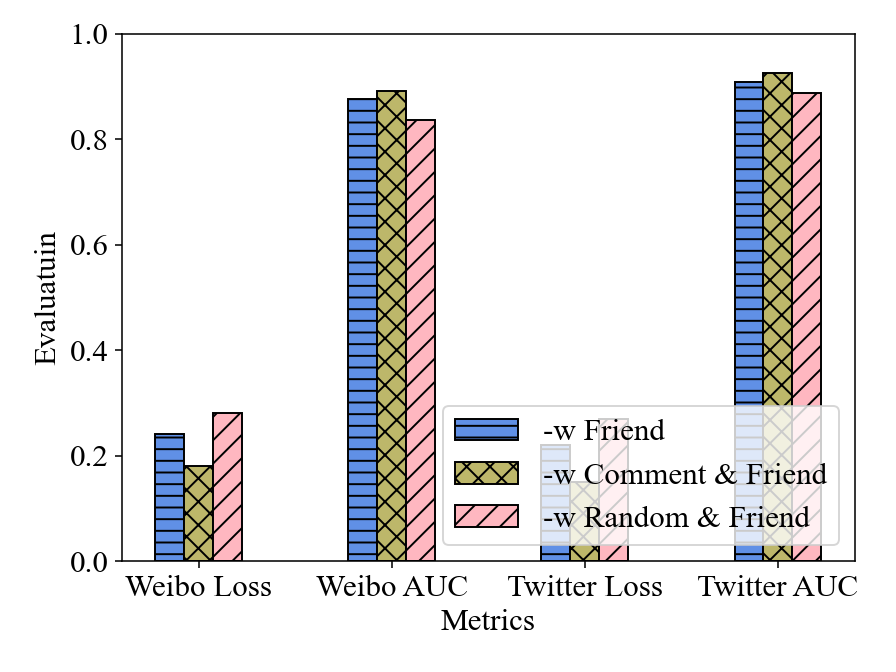}
		\end{minipage}%
	}%
	\caption{Comparison with different user relations}
	\label{fig-ur}
\end{figure}
\par As shown in Fig. \ref{fig-ur}(a), the one based on friend relations worked best. Subsequently, the model based on comment relations has the second-highest performance.
\par After that, the best combination of relations is verified based on friend relations. As shown in Fig. \ref{fig-ur}(b), the best performance is achieved by combining comment relations. When combining random relations, the model performance produces a decrease. Therefore, this section validates the best user relationship combination.
\subsection{Validity Analysis for Video Modeling Component}
\par To validate the best video modeling strategy, the experimental setup is as follows: \textbf{-w 3D-CNN}\cite{uchiyama2023visually}: using 3D-CNN components to model the video. \textbf{-w CLIP} \cite{ali2023clip}: using pre-trained CLIP components. \textbf{-w MVIT} \cite{li2022mvitv2}: using pre-trained vision Transformer (ViT) components.
\par
\begin{figure}[H]	
	\centering 
	\subfigure[WEIBO]{
		\begin{minipage}[t]{0.5\linewidth}
			\centering
			\includegraphics[width=1\linewidth]{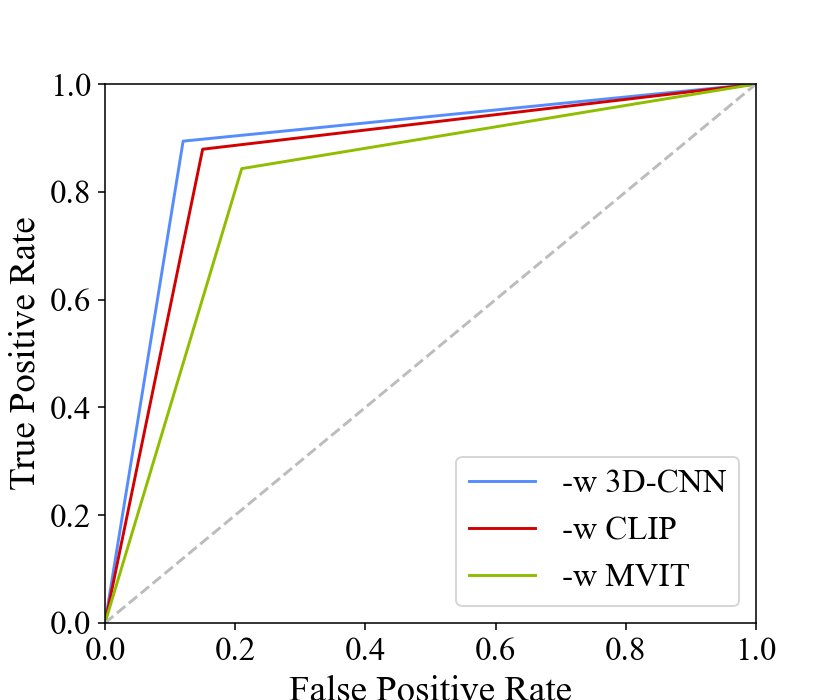}
		\end{minipage}%
	}%
	\subfigure[TWITTER]{
		\begin{minipage}[t]{0.5\linewidth}
			\centering
			\includegraphics[width=1\linewidth]{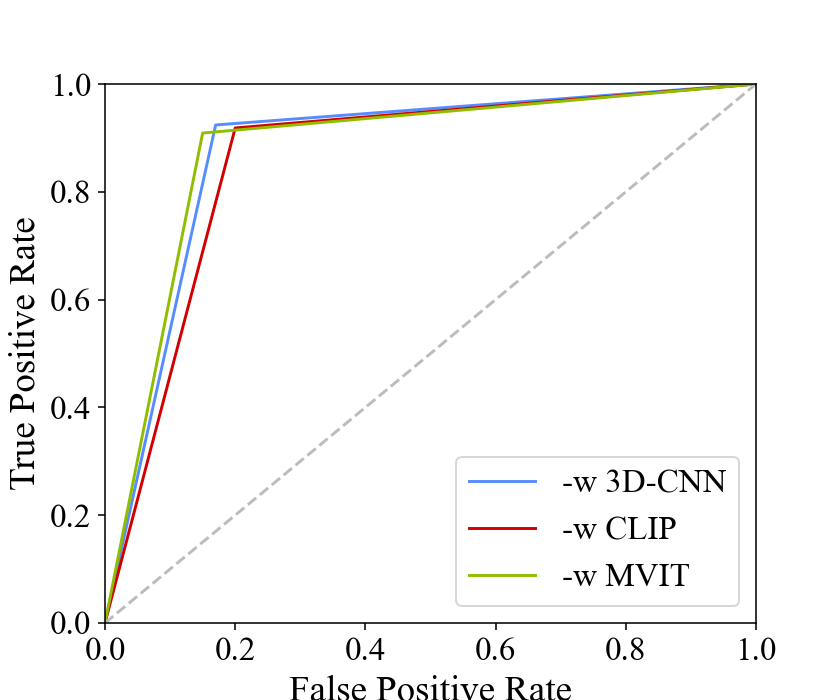}
		\end{minipage}%
	}%
	\caption{Comparison with different video modelings}
	\label{fig-u}
\end{figure}

 Comparing the pre-training performance on Weibo data (see Fig. \ref{fig-u}), CNN performs better, and CLIP and MVIT perform poorly. The reason is that the Weibo data has fewer samples. The pre-trained CLIP and MVIT models require a large number of training samples. Similarly, the MVIT component reaches number one in the Twitter data training environment. Meanwhile, the CNN component is comparable to the best performance. Therefore, this section validates the validity of using the CNN network as a video modeling component.

\section{Conclusions}
\par Inspired by video classification technology, a spammer detection model based on user videoization is proposed. The model’s performance is validated by two public datasets, WEIBO and TWITTER. Meanwhile, the model has advantages in recognition accuracy. The model still has advantages over the state-of-the-art models. In the future, the user videoization process will be further optimized to enhance the model's generalization.

\bibliographystyle{IEEEbib}
\bibliography{refs}

\begin{thebibliography}{10}

\bibitem{traganitis2021identifying}
Panagiotis~A Traganitis and Georgios~B Giannakis,
\newblock ``Identifying spammers to boost crowdsourced classification,''
\newblock in {\em ICASSP 2021-2021 IEEE International Conference on Acoustics,
  Speech and Signal Processing}, 2021, pp. 2840--2844.

\bibitem{Deng2023Markov}
Leyan Deng, Chenwang Wu, Defu Lian, Yongji Wu, and Enhong Chen,
\newblock ``Markov-driven graph convolutional networks for social spammer
  detection,''
\newblock {\em IEEE Transactions on Knowledge and Data Engineering}, vol. 35,
  no. 12, pp. 12310--12322, 2023.

\bibitem{hu2024f2gnn}
Guanghui Hu, Yang Liu, Qing He, and Xiang Ao,
\newblock ``F2gnn: An adaptive filter with feature segmentation for graph-based
  fraud detection,''
\newblock in {\em ICASSP 2024-2024 IEEE International Conference on Acoustics,
  Speech and Signal Processing}, 2024, pp. 6335--6339.

\bibitem{wu2020graph}
Yongji Wu, Defu Lian, Yiheng Xu, Le~Wu, and Enhong Chen,
\newblock ``Graph convolutional networks with markov random field reasoning for
  social spammer detection,''
\newblock in {\em Proceedings of the AAAI Conference on Artificial
  Intelligence}, 2020, vol.~34, pp. 1054--1061.

\bibitem{yi2023spammer}
Muyang Yi, Dong Liang, Rui Wang, Yue Ding, and Hongtao Lu,
\newblock ``Spammer detection on short video applications: A new challenge and
  baselines,''
\newblock in {\em ICASSP 2023-2023 IEEE International Conference on Acoustics,
  Speech and Signal Processing}, 2023, pp. 1--5.

\bibitem{wu2024graph}
Yang Wu, Jing Yang, Liming Wang, and Zhen Xu,
\newblock ``Graph-aware multi-view fusion for rumor detection on social
  media,''
\newblock in {\em ICASSP 2024-2024 IEEE International Conference on Acoustics,
  Speech and Signal Processing}, 2024, pp. 9961--9965.

\bibitem{liu2024disentangled}
Haoyu Liu, Yuanhai Xue, and Xiaoming Yu,
\newblock ``Disentangled graph representation with contrastive learning for
  rumor detection,''
\newblock in {\em ICASSP 2024-2024 IEEE International Conference on Acoustics,
  Speech and Signal Processing}, 2024, pp. 6470--6474.

\bibitem{Xiao2022Diffusion}
Yunpeng Xiao, Zhen Huang, Qian Li, Xingyu Lu, and Tun Li,
\newblock ``Diffusion pixelation: A game diffusion model of rumor \& anti-rumor
  inspired by image restoration,''
\newblock {\em IEEE Transactions on Knowledge and Data Engineering}, 2022.

\bibitem{Pang2023Topic}
Yucai Pang, Xuehong Li, Shihong Wei, Qian Li, and Yunpeng Xiao,
\newblock ``Topic to image: A rumor detection method inspired by image forgery
  recognition technology,''
\newblock {\em IEEE Transactions on Computational Social Systems}, 2023.

\bibitem{agarwal2022modeling}
Prabhat Agarwal, Manisha Srivastava, Vishwakarma Singh, and Charles Rosenberg,
\newblock ``Modeling user behavior with interaction networks for spam
  detection,''
\newblock in {\em Proceedings of the 45th International ACM SIGIR Conference on
  Research and Development in Information Retrieval}, 2022, pp. 2437--2442.

\bibitem{ZHANG2023Detecting}
Fuzhi Zhang, Jiayi Wu, Peng Zhang, Ru~Ma, and Hongtao Yu,
\newblock ``Detecting collusive spammers with heterogeneous graph attention
  network,''
\newblock {\em Information Processing \& Management}, vol. 60, no. 3, pp.
  103282, 2023.

\bibitem{yang2012analyzing}
Chao Yang, Robert Harkreader, Jialong Zhang, Seungwon Shin, and Guofei Gu,
\newblock ``Analyzing spammers' social networks for fun and profit: a case
  study of cyber criminal ecosystem on twitter,''
\newblock in {\em Proceedings of the 21st International Conference on World
  Wide Web}, 2012, pp. 71--80.

\bibitem{Song2021CED}
Changhe Song, Cheng Yang, Huimin Chen, Cunchao Tu, Zhiyuan Liu, and Maosong
  Sun,
\newblock ``Ced: Credible early detection of social media rumors,''
\newblock {\em IEEE Transactions on Knowledge and Data Engineering}, vol. 33,
  no. 8, pp. 3035--3047, 2021.

\bibitem{ma2016detecting}
Jing Ma, Wei Gao, Prasenjit Mitra, Sejeong Kwon, Bernard~J Jansen, Kam-Fai
  Wong, and Meeyoung Cha,
\newblock ``Detecting rumors from microblogs with recurrent neural networks,''
\newblock in {\em Proceedings of the International Joint Conference on
  Artificial Intelligence}, 2016, pp. 3818--3824.

\bibitem{Yang2024model}
Zhou Yang, Yucai Pang, Qian Li, Shihong Wei, Rong Wang, and Yunpeng Xiao,
\newblock ``A model for early rumor detection base on topic-derived domain
  compensation and multi-user association,''
\newblock {\em Expert Systems with Applications}, vol. 250, pp. 123951, 2024.

\bibitem{xu2023contrastive}
Yingrui Xu, Jingyuan Hu, Jingguo Ge, Yulei Wu, Tong Li, and Hui Li,
\newblock ``Contrastive learning at the relation and event level for rumor
  detection,''
\newblock in {\em ICASSP 2023-2023 IEEE International Conference on Acoustics,
  Speech and Signal Processing}, 2023, pp. 1--5.

\bibitem{zhu2012discovering}
Yin Zhu, Xiao Wang, Erheng Zhong, Nathan Liu, He~Li, and Qiang Yang,
\newblock ``Discovering spammers in social networks,''
\newblock in {\em Proceedings of the AAAI Conference on Artificial
  Intelligence}, 2012, pp. 171--177.

\bibitem{hu2014social}
Xia Hu, Jiliang Tang, Huiji Gao, and Huan Liu,
\newblock ``Social spammer detection with sentiment information,''
\newblock in {\em 2014 IEEE International Conference on Data Mining}, 2014, pp.
  180--189.

\bibitem{li2018ssdmv}
Chaozhuo Li, Senzhang Wang, Lifang He, S~Yu Philip, Yanbo Liang, and Zhoujun
  Li,
\newblock ``Ssdmv: Semi-supervised deep social spammer detection by multi-view
  data fusion,''
\newblock in {\em 2018 IEEE International Conference on Data Mining}, 2018, pp.
  247--256.

\bibitem{dou2020robust}
Yingtong Dou, Guixiang Ma, Philip~S Yu, and Sihong Xie,
\newblock ``Robust spammer detection by nash reinforcement learning,''
\newblock in {\em Proceedings of the 26th ACM SIGKDD International Conference
  on Knowledge Discovery \& Data Mining}, 2020, pp. 924--933.

\bibitem{Agarwal2022Model}
Prabhat Agarwal, Manisha Srivastava, Vishwakarma Singh, and Charles Rosenberg,
\newblock ``Modeling user behavior with interaction networks for spam
  detection,''
\newblock in {\em Proceedings of the 45th International ACM SIGIR Conference on
  Research and Development in Information Retrieval}, 2022, p. 2437–2442.

\bibitem{Ma2023Improving}
Jing Ma, Jun Li, Wei Gao, Yang Yang, and Kam-Fai Wong,
\newblock ``Improving rumor detection by promoting information campaigns with
  transformer-based generative adversarial learning,''
\newblock {\em IEEE Transactions on Knowledge and Data Engineering}, vol. 35,
  no. 3, pp. 2657--2670, 2023.

\bibitem{uchiyama2023visually}
Tomoki Uchiyama, Naoya Sogi, Koichiro Niinuma, and Kazuhiro Fukui,
\newblock ``Visually explaining 3d-cnn predictions for video classification
  with an adaptive occlusion sensitivity analysis,''
\newblock in {\em Proceedings of the IEEE/CVF Winter Conference on Applications
  of Computer Vision}, 2023, pp. 1513--1522.

\bibitem{ali2023clip}
Muhammad Ali and Salman Khan,
\newblock ``Clip-decoder: Zeroshot multilabel classification using multimodal
  clip aligned representations,''
\newblock in {\em Proceedings of the IEEE/CVF International Conference on
  Computer Vision}, 2023, pp. 4675--4679.

\bibitem{li2022mvitv2}
Yanghao Li, Chao-Yuan Wu, Haoqi Fan, Karttikeya Mangalam, Bo~Xiong, Jitendra
  Malik, and Christoph Feichtenhofer,
\newblock ``Mvitv2: Improved multiscale vision transformers for classification
  and detection,''
\newblock in {\em Proceedings of the IEEE/CVF Conference on Computer Vision and
  Pattern Recognition}, 2022, pp. 4804--4814.

\end{thebibliography}
\end{document}